\documentclass[journal]{IEEEtran}

\usepackage[justification=centering]{caption}
\usepackage{ amssymb }

\usepackage{cite}
\usepackage[pdftex]{graphicx}
\usepackage{amsmath, amsthm}
\usepackage[algo2e,ruled,vlined, linesnumbered]{algorithm2e}
\usepackage{newtxmath, newtxtext}
\usepackage{algorithm, algorithmic}
\usepackage{array}
\usepackage[caption=false,font=footnotesize]{subfig}
\usepackage{url}
\usepackage{hyperref}
\usepackage{enumitem}
\usepackage{color}
\usepackage{soul}
\usepackage{lineno}
\usepackage{listings}
\usepackage{booktabs}
\usepackage{multirow}
\usepackage[table]{xcolor}
\usepackage{subfig}
\usepackage{threeparttable}
\usepackage{doi}

\newlength\myindent
\usepackage{color}
\DeclareRobustCommand{\hlyel}[1]{{\sethlcolor{white}\hl{#1}}}

\graphicspath{{./images/}}
\DeclareGraphicsExtensions{.pdf,.jpeg,.png}

\begin{document}

\title{Coverage Path Planning for Robotic Quality Inspection with Control on Measurement Uncertainty}

\author{Yinhua Liu,~\IEEEmembership{}
        Wenzheng Zhao,~\IEEEmembership{}
        Hongpeng Liu,~\IEEEmembership{}
        Yinan Wang,~\IEEEmembership{}
        Xiaowei Yue,~\IEEEmembership{Senior Member,~IEEE}

\thanks{Y. Liu, W. Zhao and H. Liu are with the School of Mechanical Engineering, University of Shanghai for Science and Technology, Shanghai.
}
\thanks{Y. Wang and X. Yue are with the Department of Industrial and Systems Engineering, Virginia Tech, Blacksburg, VA, 24061.
}
\thanks{\textit{(Corresponding author: Xiaowei Yue, e-mail: xwy@vt.edu)}}
\thanks{\copyright 2022 IEEE. Personal use of this material is permitted. Permission from IEEE must be obtained for all other uses, in any current or future media, including reprinting/republishing this material for advertising or promotional purposes, creating new collective works, for resale or redistribution to servers or lists, or reuse of any copyrighted component of this work in other works.}
}

\markboth{ACCEPTED BY IEEE/ASME TRANSACTIONS ON MECHATRONICS.}{}

\maketitle

\begin{abstract}
The optical scanning gauges mounted on the robots are commonly used in quality inspection, such as verifying the dimensional specification of sheet structures. Coverage path planning (CPP) significantly influences the accuracy and efficiency of robotic quality inspection. Traditional CPP strategies focus on minimizing the number of viewpoints or traveling distance of robots under the condition of full coverage inspection. The measurement uncertainty when collecting the scanning data is less considered in the free-form surface inspection. To address this problem, a novel CPP method with the optimal viewpoint sampling strategy is proposed to incorporate the measurement uncertainty of key measurement points (MPs) into free-form surface inspection. At first, the feasible ranges of measurement uncertainty are calculated based on the tolerance specifications of the MPs. The initial feasible viewpoint set is generated considering the measurement uncertainty and the visibility of MPs. Then, the inspection cost function is built to evaluate the number of selected viewpoints and the average measurement uncertainty in the field of views (FOVs) of all the selected viewpoints. Afterward, an enhanced rapidly-exploring random tree (RRT*) algorithm is proposed for viewpoint sampling using the inspection cost function and CPP optimization. Case studies, including simulation tests and inspection experiments, have been conducted to evaluate the effectiveness of the proposed method. Results show that the scanning precision of key MPs is significantly improved compared with the benchmark method. 
\end{abstract}


\IEEEpeerreviewmaketitle

\section{Introduction}
\IEEEPARstart{F}{ree-form} sheet metal products, such as autobody, aircraft fuselage, and high-speed train body, have complex structures. These products have strict geometric dimensions and tolerance specifications in the forming and multistage assembly processes. Therefore, comprehensive and accurate inspection of the free-form sheet metal parts not only evaluates their manufacturing accuracy but also provides an essential basis to ensure the quality of an assembled product. In recent years, automated robotic inspection systems that are embedded with high-resolution and rapid optical sensors, can provide effective non-contact measurement. Therefore, they have been increasingly used in verifying dimensional and geometrical specifications when assembling various products, including autobody, aircraft, etc. Optical scanners can obtain a large number of point cloud data in a short time. These data capture the sizes, positions, and shapes of products and are ultimately used to calculate dimension errors. Besides, the non-contact measurement avoids damages during the inspection. 

During the quality inspection of free-form surfaces, a robotic inspection system is usually used to scan the target surface thoroughly in different viewpoints and poses. The planning of an inspection path, in essence, is formulated as the coverage path planning (CPP) problem. The CPP is defined as planning an optimal path to make the robot cover all the key measurement points (MPs) in the target surface and avoid all obstacles. Next, we review methods for solving CPP problems.

\subsection{Literature Review}
CPP has been integrated into many robotic applications, such as autonomous underwater vehicles, inspection robots, cleaning robots, etc. Galceran and Carreras did a survey on CPP for robotics \cite{galceran2013survey}. The existing CPP methods can be summarized into two categories. 

We first review the CPP methods designed in a two-dimensional (2D) space. Many scholars have done many works to solve CPP problems in 2D space, such as bridge reconstruction and mobile robot cleaning \cite{song2018varepsilon, an2018triangulation,an2017rainbow}. A mechatronic system was designed for autonomous robotic inspection and evaluated on the bridge deck inspection \cite{la2013mechatronic}. It used trapezoidal decomposition to solve the CPP problem and controlled the robot to follow the trajectory precisely. Lim et al. proposed a robotic crack inspection and mapping system \cite{lim2014robotic}, where the CPP problem is solved by the genetic algorithm. The proposed method could find a solution to minimize the number of turns and traveling distance and realize the efficient bridge deck inspection.    

In a three-dimensional (3D) space, the robot may not obtain the information of the occluded parts through a single viewpoint, which makes the CPP methods designed for the 2D space inapplicable. 3D CPP methods are specifically designed to implement robotic inspection in a 3D space \cite{yazici2013dynamic, mavrinac2014semiautomatic, olivieri2014coverage, papadopoulos2013asymptotically}. Jensen-Nau et al. leveraged the Voronoi diagrams to generate a field for near-optimal paths of viewpoints and apply constraints on path length constraints \cite{jensen2020near}. This approach can solve the CPP problem with energy constraints. Wang et al. investigated joint path planning and temporal scheduling for airborne surveillance problems and developed a multi-objective evolutionary algorithm \cite{wang2018multiperiod}. Hassan and Liu developed an adaptive CPP approach that can quickly respond to the unexpected changes in the inspected region and achieve complete coverage \cite{hassan2019ppcpp}. 

For the 3D CPP problem in product quality inspection, current studies mainly focus on obtaining the complete surface topography with the minimum inspection cost. Recently, the proposed methods generated the optimal inspection path by ensuring full coverage and minimizing the number of viewpoints. For example, Rafaeli et al. classified each measurement element based on the distance and normal direction of the surface \cite{raffaeli2013off}. In each classified set, a viewpoint was randomly selected to cover the elements in the same set to ensure the full coverage of the path. A CPP inspired by computer-aided manufacturing was developed for automated structural inspection \cite{macleod2016machining}. Glorieux et al. established a target sampling strategy based on the inspection path and the number of MPs, and then, the best viewpoints are iteratively generated \cite{glorieux2020coverage}. This strategy significantly reduces the inspection time. 

\subsection{Motivation of Incorporating Measurement Uncertainty in CPP}
The objective of current CPP methods is to obtain the overall dimension of a free-form surface accurately and efficiently. However, in real practice, accurately measuring the shape of the key MPs, such as holes, slots, trimmings, etc., in the free-form surface are rather concerned. The dimensions of these MPs directly influence the quality of final products. Furthermore, the measurement uncertainty inevitably exists in the scanning system and decays its measurement accuracy, which is mainly influenced by the optical parameters of a scanner. The pitch angle, deflection angle, and depth of field (DOF) are also critical factors in determining the measurement uncertainty \cite{gerbino2016influence}. Among these factors, the incident angle of the scanner and the measurement uncertainty are most closely related. Recent researchers tried to select the viewpoints with optimal incident angles to reduce the measurement uncertainty and improve the quality of collected data in the metrology domain. Given the dimension tolerances of key MPs, Mahmud et al. built a model to inversely infer the feasible range of measurement uncertainty to meet the tolerances \cite{mahmud20113d}. The viewpoint evaluation function has been built based on the relationship between measurement uncertainties and different incident angles, and it is further used to evaluate and select the optimal viewpoints\cite {fan2016automated}. However, it still lacks a method to incorporate the influence of various measurement uncertainties into solving the CPP problem. Measurement uncertainties not only impact the inspection accuracy and quality evaluation but also have a significant influence on the subsequent predictive modeling and quality control \cite{yue2018surrogate,lee2020neural}. Therefore, it is urgently needed to develop a new CPP algorithm for robotic inspection systems to inspect key MPs by incorporating the measurement uncertainties.

\subsection{Contributions of This Paper}
In the state-of-the-art studies, the influence of optical parameters, such as incident angle and DOF, on the overall measurement accuracy was analyzed. However, the geometric dimensioning and tolerancing specifications change over different MPs on the free-form surfaces, which means that feasible ranges of measurement uncertainties for different MPs are different. For example, the holes, slots, and other key MPs that affect the assembly accuracy in the upcoming stations have higher precision requirements (require lower measurement uncertainties) than the non-critical MPs. If all MPs on the surfaces are inspected with the same level of measurement uncertainty, the measurement accuracy of some key MPs may not meet their quality tolerances. Meanwhile, the quality inspection may be inefficient. Therefore, we propose a novel CPP method for robotic quality inspection by incorporating the requirements of various measurement uncertainties that are determined by the corresponding tolerancing specifications. Specific contributions include: 

\begin{itemize}
    
    \item[1)] The proposed CPP method focuses on accurately and efficiently measuring the key MPs with various requirements of measurement uncertainties. To the best of our knowledge, it is the first time that measurement uncertainties have been incorporated into solving the CPP problem. The optical parameters in measuring each MP are determined separately to meet its own requirement of measurement uncertainty.
    \item[2)] The enhanced rapidly-exploring random tree (RRT*) algorithm is specifically designed to generate the optimal subset of viewpoints so as to ensure the visibility and full coverage of all MPs, reduce mean measurement uncertainty of each viewpoint, and improve the efficiency of surface inspection.
    \item[3)] The scanning path planning algorithm is developed to generate the best collision-free path that minimizes the overall inspection time and ensures each MP is visited exactly once.
\end{itemize}

The remainder of this paper is organized as follows. In Section \ref{sec3}, the flow chart is firstly introduced to demonstrate the big picture of the proposed CPP method, and then, the details are discussed step by step. In Section \ref{sec4}, an experiment on sheet metal parts is conducted to verify the effectiveness of the proposed method. The conclusions are summarized in Section \ref{sec5}.

\section{CPP incorporating measurement uncertainty}\label{sec3}

\begin{figure}[ht]
    \includegraphics[width=\linewidth]{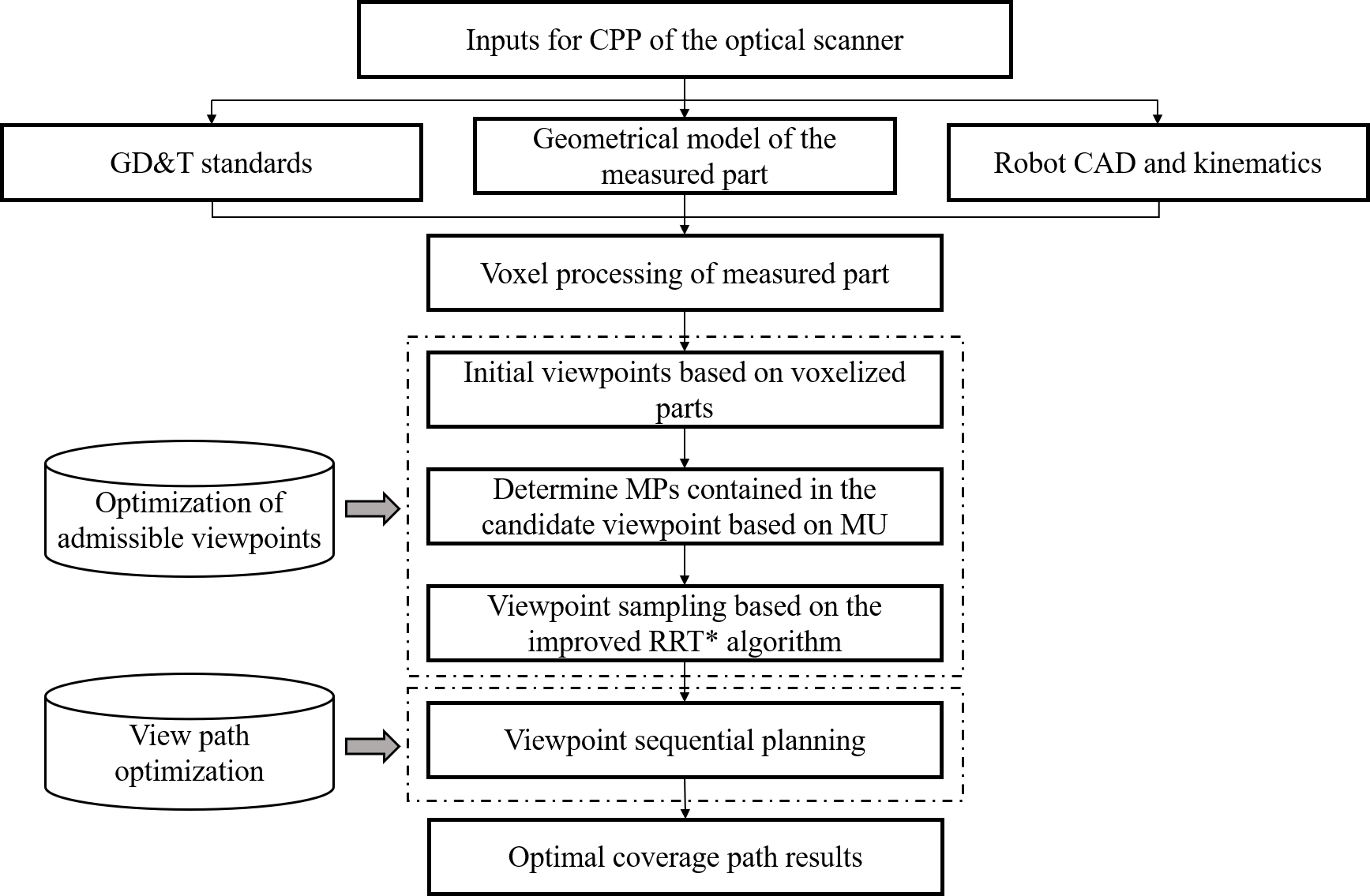}
    \centering
    \caption{Flow chart of the proposed CPP method}
    \label{Fig1}
\end{figure}

The CPP problem can be decomposed into two sub-problems: 1) Viewpoint sampling is to generate the least number of viewpoints to simultaneously achieve the full coverage of the inspected part and meet the requirements of the measurement uncertainties. 2) Path planning is to find an optimal or near-optimal sequence of all viewpoints and, at the same time, generate a collision-free inspection path. Fig. \ref{Fig1} shows the flowchart of the proposed CPP method in this paper. 

\subsection{Influencing Factors of Measurement System Uncertainty}
\label{sec3.1}
For the sheet metal parts inspection, the key MPs (e.g., holes, slots, trimmings, threads, and surface points) and their corresponding tolerances are defined according to the engineering design. The structured light scanner collects the point cloud data of the entire surface. The dimensions of the key MPs are extracted based on the point cloud data. The measurement uncertainty represents the accuracy of the collected point cloud data, and it is mainly influenced by random errors in the inspection system. To satisfy the required tolerance ($T$) of a specific MP \cite{sadaoui2019computer}, Mahmud et al. proposed a criterion to validate the capabilities of a laser-plane sensor when performing 3D metrology \cite{mahmud20113d}. Based on ISO14253 standard \cite{iso2008uncertainty}, the following rule was proposed to link the expanded uncertainty ($U$) to the tolerance interval ($T$) of the geometrical specification.


\begin{align}
    U \leq \frac{T}{8}.
    \label{eq 3-1}
\end{align}

For the inspection system, the uncertainty is from the entire measurement process. It mainly consists of method uncertainty and realization uncertainty\cite{iso2008uncertainty}. In this problem, we mainly consider the random uncertainty of the sensor $U_{sen}$, the uncertainty of the material $U_{mat}$ and the uncertainty of the robot $U_{rot}$. According to the industrial standards of ISO-14253 \cite{iso2008uncertainty}, the combined standard uncertainty $U_{ad}$ can be expressed as

\begin{align}
    U_{ad}=\sqrt{U_{sen}^{2}+U_{mat}^{2}+U_{rot}^{2}}.
    \label{eq 3-2}
\end{align}

The expanded uncertainty $U$ can be further expressed as the multiple of the combined standard uncertainty $U_{ad}$, which is $U = kU_{ad}$ \cite{iso2008uncertainty}. To this point, Equation (\ref{eq 3-1}) can be reformulated as

\begin{align}
    \frac{U}{T}=\frac{kU_{ad}}{T} \leq \frac{1}{8},
    \label{eq 3-20}
\end{align}
where $k$ is the coverage factor, which generally takes $1,2,3$. For a normal distribution, when $k=2$, the confidence level of the measurement uncertainty is $95\%$.
\begin{figure}[ht]
    \includegraphics[width=0.8\linewidth]{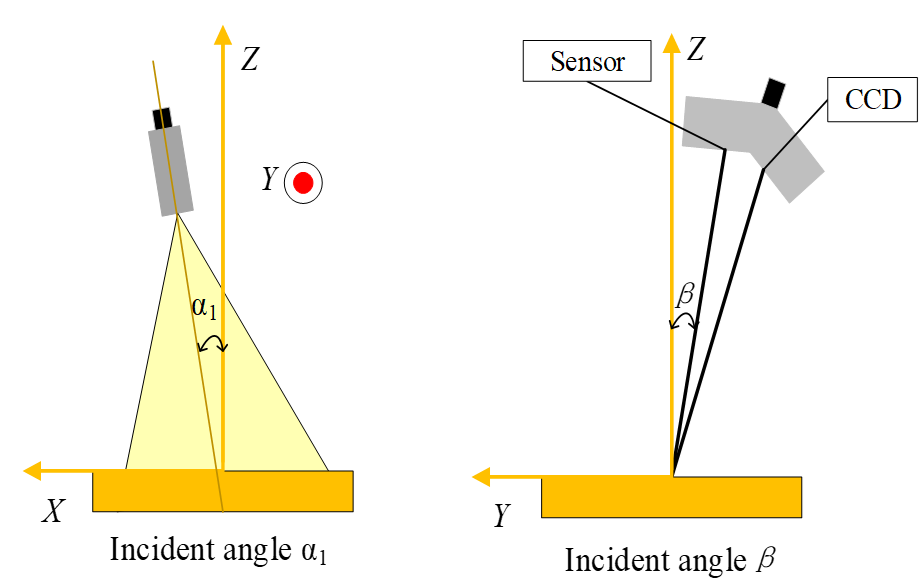}
    \centering
    \caption{Positioning and orientation parameters of the scanner.}
    \label{Fig20}
\end{figure}

\begin{figure}[ht]
    \includegraphics[width=50mm]{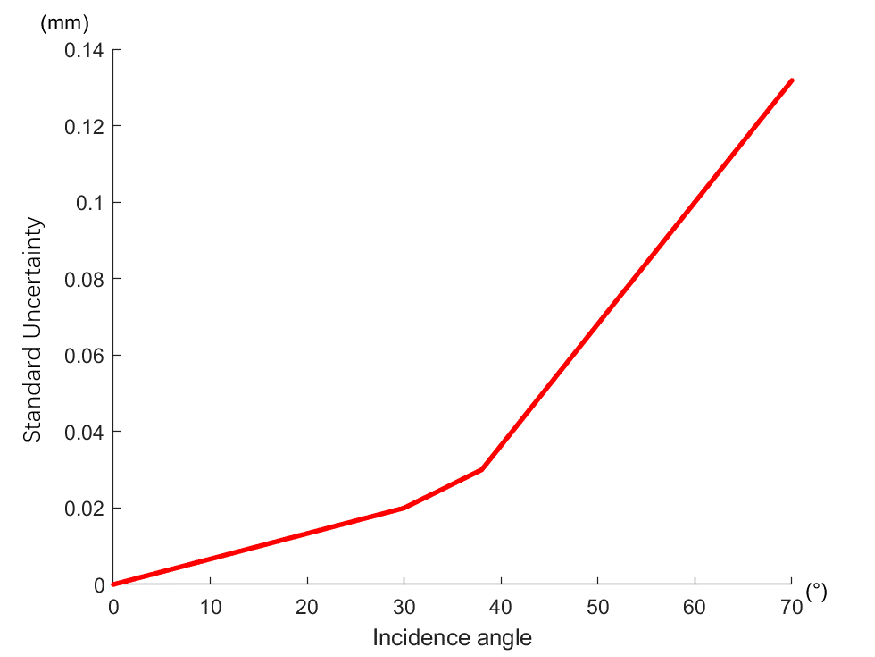}
    \centering
    \caption{The effect of the incident angle on sensor uncertainty.}
    \label{Fig2}
\end{figure}

The sensor uncertainty $U_{sen}$ is the most significant one. It is influenced by many factors, which makes the evaluation of sensor uncertainty very complicated. Many research works tried to simplify the evaluation of sensor uncertainty. Isheil et al.\cite{isheil2011systematic}  found that the incident angles $\alpha_1$ and $\beta$ have a greater influence on the sensor uncertainty, compared with the relative positioning ${d}$. As shown in Fig. 2, $\alpha_1$ is the angle of incidence (in the scanning plane) between the axis of the laser plane and the normal axis of the measured surface. $\beta$ is the angle of ortho-incidence (in the plane orthogonal to the scanning plane) between the axis of the laser plane and the normal axis of the measured surface. Mahmud et al. used laser beams instead of laser planes to simplify the visual model of the sensor \cite{mahmud20113d}. The sensor uncertainty mainly comes from the random errors of laser beams. It is associated with the laser incidence angle $\alpha$, which is between the symmetry axis of the laser plane and the normal vector of the measured surface. The angles $\alpha_1$ and $\beta$ are the projection of $\alpha$ on the laser plane and on a frontal plane, respectively. Under a given relative positioning ${d}$, the relationship between the incident angle $\alpha$ and the sensor uncertainty $U_{sen}$ can be determined through experiments \cite{Soudarissanane}. Therefore, the sensor uncertainty can be expressed as a function of the incident angle $\alpha$, i.e., $U_{sen}=f(\alpha)$. The influence of different materials and robots on $f(\alpha)$ is neglected. Fig. \ref{Fig2} shows the relationship between the incident angle and the sensor uncertainty.

The material uncertainty $U_{mat}$ is mainly from surface roughness, specular reflections, transparency, etc., which can be significantly reduced by preprocessing the material, such as spraying. We assume that the sheet metal parts to be inspected have been well preprocessed in this study. The robot uncertainty $U_{rot}$ can be estimated based on the calibration. Therefore, the influence of $U_{mat}$ and $U_{rot}$ can be kept subtle. The feasible range of sensor uncertainty ($U_{sen}$) when measuring a specific MP can be obtained by solving Equations (\ref{eq 3-1}) and (\ref{eq 3-2}), which is given in Equation (\ref{eq 3-13}). With the relationship between $U_{sen}$ and the incident angle $\alpha$, the feasible range of incident angle for a specific MP can be calculated. When $\alpha$ is in this range, the measurement uncertainty of the MP can satisfy the tolerance specification. Furthermore, the measurement uncertainties of all the MPs associated with a viewpoint can be calculated, which will be used for the initial viewpoint sampling in the next Section.

\begin{align}
    U_{sen}\leq\sqrt{(\frac{T}{8k})^{2}-U_{mat}^{2}-U_{rot}^{2}}.
    \label{eq 3-13}
\end{align}

\subsection{Initial Viewpoints Generation Based On Measurement Specification}
\label{sec3.2}

\begin{figure}[ht]
\centering
    \includegraphics[width=\linewidth]{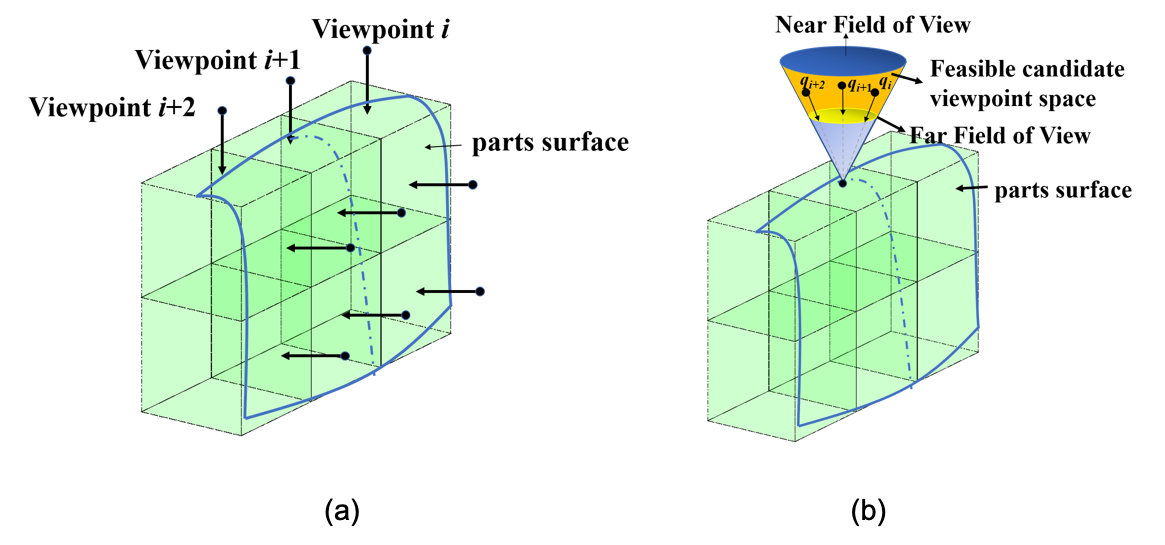}
    \caption{Schematic diagram of viewpoints initialization (a) traditional method; (b) method based on uncertainty}
    \label{Fig3}
\end{figure}

Before planning the inspection path, the part needs to be voxelized to ensure full coverage. After voxelization, each voxel contains a single key MP. In general, when the CAD model is voxelized, the following factors need to be considered: (1) The maximum length of the voxel must be less than $50\%$ of the sensor laser beam; (2) The maximum length of facets and the chordal error between the voxel and the facets should be small. Therefore, we want to ensure that the voxelized mesh model is close to the surface \cite{zhao2012automated}. As shown in Fig. \ref{Fig3} (a), the traditional method to generate viewpoints is to calculate the average normal vector of each voxel and then move the center of the voxel by a certain distance along with the opposite direction to obtain initial viewpoints.

The traditional method can reduce the number of viewpoints, but does not consider the full coverage of key MPs and their measurement uncertainties. The measurement uncertainties need to be carefully controlled to ensure the key MPs meet their required tolerances. Based on Section \ref{sec3.1}, the measurement accuracy changes with the incident angle. Therefore, when inspecting a voxel, to ensure the measurement accuracy meet the required tolerance $T$, the feasible range of incident angle $\alpha$ is calculated by Equation (\ref{eq 3-13}) and the relationship between $U_{sen}$ and $\alpha$.

\begin{figure}[ht]
    \includegraphics[width=0.6\linewidth]{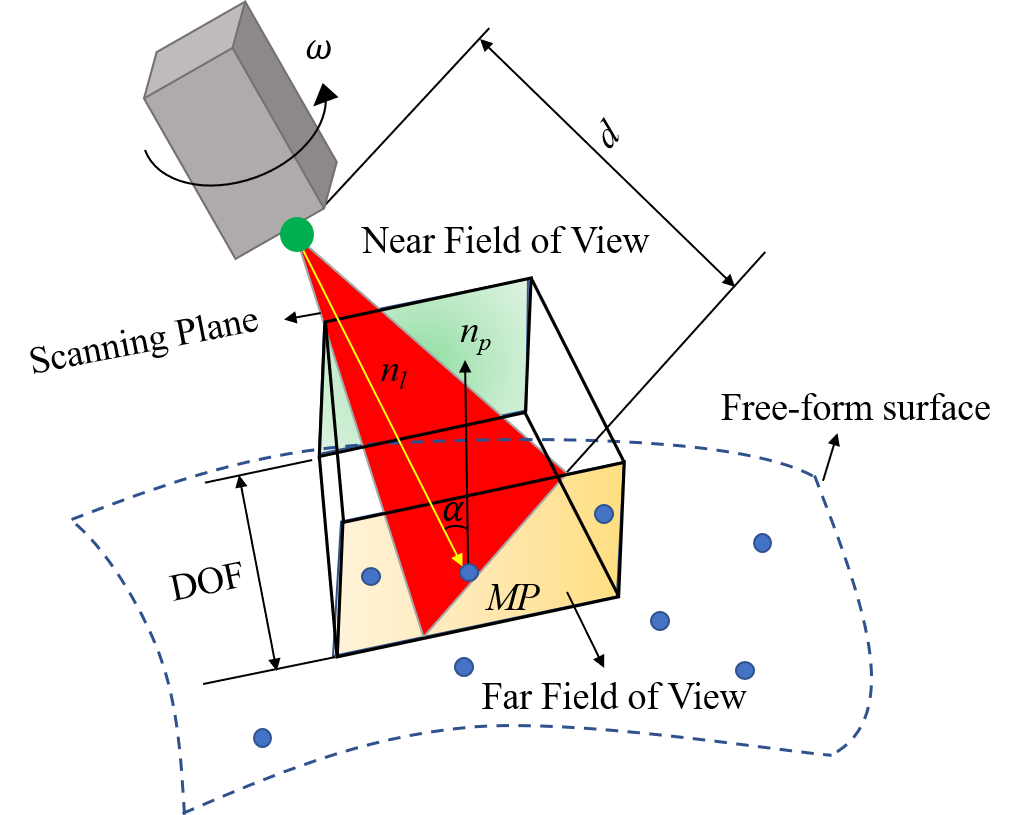}
    \centering
    \caption{Schematic diagram of the optical sensor inspection}
   \label{Fig4}
\end{figure}

The visibility of voxels to the sensor is jointly determined by the sensor's FOV and DOF. If only the sensor's FOV is considered, the feasible space of the candidate viewpoints to inspect an arbitrary voxel is a cone, which is shown in Fig. \ref{Fig3} (b). Due to the constraint of the sensor's DOF, some voxels cannot be inspected even though they are in the sensor's FOV. Therefore, it is important to further modify the feasible space of candidate viewpoints given the sensor's DOF. Fig. \ref{Fig4} shows a schematic diagram of sensor inspection, where the FOV of an optical sensor is a quadrangular prism. Given the incident angle $\alpha$ and scanning depth $d$, the sensor rotates around the axis in a certain degree $\omega$, then the number of visible key MPs in the FOV changes. Therefore, initializing the feasible space of candidate viewpoints needs to consider the following three factors: (a) feasible incident angle $\alpha$; (b) scanning depth $d$; (c) rotation angle $\omega$ of the sensor. Specific steps are as follows:

\begin{enumerate}
    \item Extract the position of the center point of the voxel and the information of key MPs contained in the voxel (such as holes, square grooves, surface points, etc.).
    \item Calculate the feasible range of incident angle $\alpha$ of each voxel center based on Equation (\ref{eq 3-13}) and the relationship between $U_{sen}$ and $\alpha$; Determine the scanning depth $d$ based on the sensor-related parameters; Determine the range of the rotation angle $\omega$ through the accessibility analysis of the robot end-effector.
    \item The initial pose of the viewpoint is determined within the feasible ranges of the incident angle $\alpha$, the scanning depth $d$, and the rotation angle $\omega$.
\end{enumerate}

Based on the aforementioned steps, the feasible poses of the viewpoints for all the voxels can be obtained, which provides a searching space for viewpoint optimization.

\subsection{Determine the Set of MPs Under the Viewpoint}
\label{sec3.3}
When planning the covering path for the free-form surface, the visible MPs located in the FOV, denoted by $\Omega$, need to be determined. Although some MPs are included in the FOV of viewpoint $\mathbf{q}_i$ (the viewpoints in this paper all contain position and pose information), these MP are not ``visible'' in $\mathbf{q}_i$ if the incident angle between those MPs and viewpoint $\mathbf{q}_i$ does not meet the requirements of measurement tolerancing. In other words, it is necessary to satisfy:

\begin{align*}
    \left|\operatorname{arccos}\left(\frac{\mathbf{n}_{l} \cdot \mathbf{n}_{p}}{\left|\mathbf{n}_{l}\right| \cdot\left|\mathbf{n}_{p}\right|}\right)\right| \leq \alpha_{p},
\end{align*}
where $\mathbf{n}_{l}$ is the normal vector of the scanning plane of the sensor in the viewpoint $\mathbf{q}_{i}$; $\mathbf{n}_{p}$ is the vector of the MP $p$; $\alpha_{p}$ represent the maximum feasible incidence angle of the MP $p$.

Moreover, the optical sensor is mounted at the end-effector of the robot to measure the object, so the robot's accessibility and the collision need to be considered for a determined viewpoint. For a given viewpoint $\mathbf{q}_{i}$, the MPs are visible only if the following four conditions are satisfied.

\begin{enumerate}
    \item The robot satisfies the pose accessibility to  $\mathbf{q}_i$.
    \item There is no collision between the robot and the static environment for a specified viewpoint pose. 
    \item The MP locates in the feasible FOV of $\mathbf{q}_i$.
    \item The incident angle is within the feasible range determined by the measurement uncertainties of MPs.
\end{enumerate}

Algorithm \ref{algorithm1} presents the procedure to extract the set of MPs $\mathcal{N}_{\mathbf{q}_{i}}$ and its number $n_{i}$ under the viewpoint $\mathbf{q}_{i}$. The functions used in Algorithm \ref{algorithm1} can be found in appendix A.

\begin{algorithm}[ht]
  \algsetup{linenosize=\tiny}
  \scriptsize
\SetAlgoLined
    Function: $(\mathcal{N}_{\mathbf{q}_{i}}, n_{i})=\textbf{DeterminationSet}(\mathbf{q}_{i}, \Omega, \mathbf{P})$ \\
    $n_{i}=0; \mathcal{N}_{\mathbf{q}_{i}} = \emptyset$\\
    $\mathbf{S} = \textbf{BoundingBoxSet}(\Omega, \mathbf{P})$\\
    \If{$\textbf{\upshape RobotAccessibility}(\mathbf{q}_{i})=0$}{
    $n_{i}=0; \mathcal{N}_{\mathbf{q}_{i}} = \emptyset$
    }{\If{$\textbf{\upshape Collision}(\mathbf{q}_{i})=1$}{
    $n_{i}=0; \mathcal{N}_{\mathbf{q}_{i}} = \emptyset$
    }
    {\For{$i=1:m_{\mathbf{S}}$}{\If{$\left|\operatorname{acos}\left(\frac{\mathbf{n}_{l} \cdot \mathbf{n}_{\mathbf{S}(i)}}{\left|\mathbf{n}_{l}\right| \cdot\left|\mathbf{n}_{\mathbf{S}(i)}\right|}\right)\right| \leq \alpha_{\mathbf{S}(i)}$}{$n_{i}=n_{i}+1; \mathcal{N}_{\mathbf{q}_{i}} = \mathcal{N}_{\mathbf{q}_{i}} \bigcup \mathbf{S}(i)$}}
    }
    }
    \textbf{return} $\mathcal{N}_{\mathbf{q}_{i}}, n_{i}$
\caption{Algorithm to extract the set of MPs}
\label{algorithm1}
\end{algorithm}

In Algorithm \ref{algorithm1}, $\mathbf{P}$ is the full set of MPs; $\mathbf{S}$ is the subset of MPs contained in $\Omega$; $\Omega$ is the visible MPs located in the FOV; $m_{\mathbf{S}}$ is the number of measurement points contained in $\mathbf{S}$.

\subsection{Viewpoint Sampling Via the Enhanced RRT*}
\label{sec3.4}
The initial viewpoints need to be further sampled to reduce the redundancy and improve the efficiency and precision of CPP in quality inspection. The criteria of viewpoints sampling can be summarized into two aspects.
\begin{enumerate}
    \item All the MPs in the inspected part need to be covered with the minimum number of viewpoints.
    \item The measurement uncertainties of the MPs under the viewpoint should be as small as possible to improve the scanning precision and meet the requirements of measurement accuracy.
\end{enumerate}

Therefore, we design the objective function for viewpoint sampling to meet these criteria simultaneously. More specifically, the first term in the objective function denotes the average uncertainty of the MPs that are visible in the FOV, and the second term denotes the number of viewpoints. The objective function is,

\begin{align}
     f\left(m, n_{i}^{\prime}\right)=\beta_{1} \sum_{i=1}^{m} \sum_{j=1}^{n_{i}^{\prime}} \frac{U\left(\mathcal{N}_{\mathbf{q}_{i}}^{\prime}(j)\right)}{n_{i}^{\prime}}+\gamma_{1} m.
        \label{eq 3-3}
\end{align}

Thus, the optimization problem is formulated as:
\begin{align}
   \min f\left(m, n_{i}^{\prime}\right),
   \label{eq 3-4}
   \\
    \text { s.t. } \mathcal{N}_{\mathbf{q}_{1}}^{\prime} \cup \ldots \cup \mathcal{N}_{\mathbf{q}_{i}}^{\prime} \cup \ldots \cup \mathcal{N}_{\mathbf{q}_{m}}^{\prime}=\mathbf{P},
    \label{eq 3-11}
      \\
    \sum_{i=1}^{m} n_{i}^{\prime}=N,
    \label{eq 3-8}
    \\
    \mathcal{N}_{\mathbf{q}_{i }}^{\prime} \neq \emptyset,
    \label{eq 3-5}
    \\
    \mathbf{G}\left(\mathbf{q}_{i}\right) \cap \boldsymbol{V}=\emptyset,
    \label{eq 3-6}
     \\
    n_{i }^{\prime},m \text { are both positive integers,}
    \label{eq 3-7}
\end{align}
where $m$ is the total number of selected viewpoints; $N$ is the number of voxels; $\mathcal{N}_{\mathbf{q}_{i}}^{'}$ is set of MPs newly included in viewpoint $\mathbf{q}_{i}$; $n_{i}^{'}$ is the number of MPs in set $\mathcal{N}_{q_i}^{'}$; the measurement uncertainty of the $j_{\text{th}}$ MP in $\mathcal{N}_{q_i}^{'}$ is denoted as $U(\mathcal{N}_{\mathbf{q}_{i}}^{'}(j))$; the geometric space occupied by the inspection system at the viewpoint $\mathbf{q}_{i}$ is denoted as $\mathbf{G}(\mathbf{q}_{i})$; the geometric space occupied by the inspected surface, including the inspected free-form part, tooling bracket system, etc., is denoted as $\mathbf{V}$; $\beta_{1}$ and $\gamma_{1}$ are the weights in the objective function.

The primary objective of CPP is to reduce the number of viewpoints and the changes of probe poses as much as possible to improve inspection efficiency when ensuring the full coverage of the surface. In Section \ref{sec3.2} and \ref{sec3.3}, the initial redundant viewpoints and the corresponding visible MPs under each viewpoint are determined. To reduce the number of viewpoints and meet the measurement uncertainty requirements, the inspection viewpoints are sampled by solving the model established in Equations (\ref{eq 3-3}-\ref{eq 3-7}), and the enhanced RRT* algorithm is proposed to general the optimal solution.

The RRT* algorithm is an incremental method to explore the collision-free path of high-dimension space by increasing the number of search tree nodes. The RRT* algorithm has asymptotic optimality, that is, it can converge to an optimal solution iteratively with the increase of sampling points \cite{karaman2010optimal}. The proposed enhanced RRT* algorithm generally follows three steps: First, the new candidate viewpoint $\mathbf{q}_{new}$ is determined from the viewpoint set $\mathcal{T}$ by using the proposed $\textbf{Extend}()$ function (Algorithm \ref{algorithm3}). Second, the best parent node of $\mathbf{q}_{new}$ is selected and denoted as $\mathbf{q}_{min}$ based on the cost function (Equation (\ref{eq 3-12})) and Algorithm \ref{algorithm6}. Third, all the included MPs $\mathbf{P}_{\mathbf{q}_{new}}$ can be determined by tracing back from the current root node $\mathbf{q}_{new}$ to its parent node $\mathbf{q}_{min}$ and repeating the tracing procedure recursively until reaching the initial viewpoint. The directed tree graph $\mathcal{G}$ is extended by a tuple of nodes and MPs $\{\left(\mathbf{q}_{new}, \mathbf{q}_{\min}\right), \mathbf{P}_{\mathbf{q}_{new}}\}$. The algorithm terminates when $\mathbf{P}_{\mathbf{q}_{new}}$ includes all the key MPs, or the algorithm reaches the maximum number of iterations. The last optimal viewpoints $\mathbf{q}_{best}$ and its parent node are determined as the final optimal viewpoint through the directed tree graph $\mathcal{G}$. The overview of the proposed enhanced RRT* algorithm is given in Algorithm \ref{algorithm2}. Functions used in Algorithm \ref{algorithm2} are introduced separately.

\begin{algorithm}[ht]
  \algsetup{linenosize=\tiny}
  \scriptsize
\SetAlgoLined
    $\mathcal{T} \xleftarrow{} \textbf{CalSampling}(env,\mathbf{P})$  \\
    $ \mathbf{Q} = \mathbf{q}_{init}$; $i$=$0$;  $\mathcal{G}=\emptyset$; $\mathbf{q}_{{nearest}}=\mathbf{q}_{init}$;\\
      \While{{$i$}<{$i_{max}$}}{$\mathbf{q}_{{rand }}=\textbf{Sample}(\mathcal{T})$\\
    $\mathbf{q}_{{new }}=\textbf {Extend}\left(\mathbf{q}_{{rand }}, \mathbf{q}_{{nearest }}, \mathbf{Q}, \mathcal{T}\right)$\\
    $\mathbf{q}_{{nearest }}=\textbf { FindbestNeighbor}\left(\mathbf{Q}, \mathbf{q}_{{new }}\right)$\\
   \If {$ \textbf{\upshape FeasiblePoint}(\mathbf{q}_{new})=1$}{
    $\mathbf{q}_{near}=\textbf {Neighbors}(\mathbf{Q}, \mathbf{q}_{new})$\\
    $\mathbf{q}_{min}=\textbf {ChooseParent}\left(\mathbf{q}_{{near}}, \mathbf{q}_{{new}}, \mathbf{q}_{{nearest}}\right)$\\
    $\mathbf{Q}$=$\mathbf{Q}\cup \ \mathbf{q}_{new}$\\
    
   $\mathbf{P}_{\mathbf{q}_{new}}$=$\mathbf{P}_{\mathbf{q}_{min}}\cup\mathcal{N}_{\mathbf{q}_{new}}^{\prime}$ \\

$\mathcal{G} \leftarrow\left\{\left(\mathbf{q}_{new}, \mathbf{q}_{min }\right), \mathbf{P}_{\mathbf{q}_{new}}\right\}$ \\

    }
        \If { \hlyel{$ \mathbf{P}=\mathbf{P}_{\mathbf{q}_{new}}$}}{\hlyel{
        $\mathbf{q}_{best}$=$\mathbf{q}_{new}$\\}
        $\mathbf{break}$}
    $i$=$i+1$}
    \textbf{return} $\mathcal{G}$, $\mathbf{q}_{best}$
\caption{Viewpoint determination via enhanced RRT*}
\label{algorithm2}
\end{algorithm}
In Algorithm \ref{algorithm2}, $env$ denotes the static environment including the measured part and tooling, etc.; $\mathbf{Q}$ is the selected candidate viewpoint set; $i_{max}$ is the maximum number of iterations; $\mathcal{T}$ is the candidate viewpoint set; $\mathbf{P}_{\mathbf{q}_{new}}$ is a set of MPs determined by tracing back from the current root node $\mathbf{q}_{new}$ to the initial viewpoint recursively; $\mathcal{N}_{\mathbf{q}_{new}}^{\prime}$ is the set of measured points contained in viewpoint $\mathbf{q}_{new}$; $\mathbf{q}_{best}$ is the last optimal viewpoint selected; $\mathbf{q}_{init}$ represents the initial viewpoint; $\mathbf{q}_{rand}$ represents the randomly sampled viewpoint; $\mathbf{q}_{nearest}$ represents the closest viewpoint in $\mathbf{Q}$ to $\mathbf{q}_{new}$; $\mathbf{q}_{near}$ represents viewpoint in $\mathbf{Q}$ whose distance to $\mathbf{q}_{new}$ is less than the threshold; $\mathbf{q}_{new}$ represents the newly determined viewpoint; $\mathbf{q}_{min}$ represents the viewpoint closest to $\mathbf{q}_{new}$ in $\mathcal{T}$.

The traditional RRT* algorithm determines the candidate viewpoint $\mathbf{q}_{new}$ based on a random sampling strategy. It will cause that some MPs are repetitively inspected. This will decay the measurement accuracy and efficiency. In the enhanced RRT* algorithm, the function $\textbf{Extend}()$ is designed to solve Equations (\ref{eq 3-3})-(\ref{eq 3-7}) more efficiently. The viewpoint that has the closest distance to the straight line constructed by $\mathbf{q}_{rand}$ and $\mathbf{q}_{nearest}$ is selected from the candidate viewpoint set $\mathcal{T}$. Moreover, to reduce the overlap, the distance from the selected viewpoint to $\mathbf{q}_{nearest}$ should be larger than $L$. $L$ is the width of the far FOV when the sensor is at the $\mathbf{q}_{nearest}$. When the selected viewpoint is far away from the current $\mathbf{q}_{nearest}$, although the number of unmeasured MPs in the candidate viewpoint $\mathbf{q}_{new}$ is large, some unmeasured MPs might be leftover between the selected viewpoint and $\mathbf{q}_{nearest}$. As the iteration converges, the leftover MPs are sparsely distributed over the surface. To cover them, the newly selected viewpoint might contain only a few unmeasured MPs, which leads to an increase in the number of viewpoints. To avoid this, we select the closest viewpoint whose distance to $\mathbf{q}_{nearest}$ is greater than $L$ as $\mathbf{q}_{new}$. The value of $L$ can be determined according to the width of the actual sensor FOV. The procedure of the proposed $\textbf{Extend}()$ function is shown in Algorithm \ref{algorithm3}.

\begin{algorithm}[ht]
  \algsetup{linenosize=\tiny}
  \scriptsize
\SetAlgoLined
    Function: $\mathbf{q}_{new}=\textbf{Extend}(\mathbf{q}_{rand}, \mathbf{q}_{nearest}, \mathbf{Q}, \mathcal{T})$  \\
    $\mathcal{T}_{unselect}=\mathcal{T}-\mathbf{Q}$\\
    $D_{0}=\inf$\\
    \For{$i=1:m_{un}$}{
    $D_{last} = \mathbf{Mindistance}(\mathbf{q}_{rand}, \mathbf{q}_{nearest}, \mathcal{T}_{unselect}(i))$\\
    \If{$\left|\mathcal{T}_{{unselect(i)}}-\mathbf{q}_{{nearest}}\right|>\mathrm{L}$}{
    $\mathbf{q}_{last} = \mathcal{T}_{unselect(i)}$\\
    \If{$D_{last} < D_{0}$}{
    $\mathbf{q}_{new}=\mathbf{q}_{last}$\\
    $D_{0} = D_{last}$\\
    }
    }
    }
    \textbf{return} $\mathbf{q}_{new}$
\caption{Determine new candidate viewpoint via updated Extend function}
\label{algorithm3}
\end{algorithm}

In Algorithm \ref{algorithm3}, $\textbf{Mindistance}()$ is used to calculate the shortest distance $D_{last}$ from the viewpoint locations in $\mathcal{T}_{unselect}$ to the line constructed by $\mathbf{q}_{rand}$ and $\mathbf{q}_{nearest}$; $\mathcal{T}_{unselect}$ is the viewpoint set that is not selected in $\mathcal{T}$; $m_{un}$ denotes the number of viewpoints in $\mathcal{T}_{unselect}$; $D_{0}$ denotes the shortest distance from $\mathbf{q}_{new}$ to the line constructed by $\mathbf{q}_{rand}$ and $\mathbf{q}_{nearest}$. The details are shown in Algorithm \ref{algorithm4}.

\begin{algorithm}[ht]
  \algsetup{linenosize=\tiny}
  \scriptsize
\SetAlgoLined
    Function: $D_{last}=\textbf{Mindistance}(\mathbf{q}_{rand},\mathbf{q}_{nearest}, \mathcal{T}_{unselect(i)})$\\
    $d_{1}=\left||\mathbf{q}_{{rand}}-\mathbf{q}_{{nearest}}\right||_{2}$\\
    $d_{2}=\left||\mathbf{q}_{{rand}}-T_{{unselect(i)}}\right||_{2}$\\
    $d_{3}=\left||\mathbf{q}_{{nearest}}-T_{{unselect(i)}}\right||_{2}$\\
    $s=(d_{{1}}+d_{{2}}+d_{{3}})/2$\\
    $S= \sqrt{s(s-d_{{1}})(s-d_{{2}})(s-d_{{3}})}$ \\
    $D_{last}$=$2S$/ $d_1$  \\
\textbf{return} $D_{last}$
\caption{calculate the shortest distance $D_{last}$ from the viewpoint locations in $\mathcal{T}_{unselect}$ to the line $\overline{\mathbf{q}_{rand},\mathbf{q}_{nearest}}$.}
\label{algorithm4}
\end{algorithm}

The procedure of $\textbf{FindbestNeighbor}()$ function is shown in Algorithm \ref{algorithm5}, where $m_{\mathbf{Q}}$ indicates the number of viewpoints contained in $\mathbf{Q}$; $\epsilon$ represents the distance threshold; $\textbf{Cost}$ is the modified objective function using Equation (\ref{eq 3-12}).

\begin{algorithm}[ht]
  \algsetup{linenosize=\tiny}
  \scriptsize
\SetAlgoLined
     Function: $\mathbf{q}_{nearest}=\textbf{FindbestNeighbor}(\mathbf{Q},\mathbf{q}_{new})$\\
    $cost=inf$\\
     \For{$i=1:m_{Q}$}{
     $D=\left||\mathbf{Q}(i)-\mathbf{q}_{{new}}\right||_{2}$\\
     \If{$D<\epsilon$}{
     $cost0=\textbf{Cost}(\mathbf{Q(i)},\mathbf{q}_{{new}})$\\
     \If{$cost0<cost$}{
     $\mathbf{q}_{{nearest}=\mathbf{Q(i)}}$\\
     }
     }
    } 
\textbf{return} $\mathbf{q}_{{nearest}}$ 
\caption{Search the viewpoint $\mathbf{q}_{nearest}$ that is nearest to $\mathbf{q}_{new}$.}
\label{algorithm5}
\end{algorithm}

When the candidate viewpoint $\mathbf{q}_{new}$ satisfies the robot accessibility and collision-free conditions between the robot and the environment, the parent node of $\mathbf{q}_{new}$ is determined. The traditional RRT* algorithm generates the parent node based on the distance between the candidate viewpoint $\mathbf{q}_{new}$ and $\mathbf{q}_{near}$. In this paper, the average measurement uncertainty and the number of MPs that are measured by the candidate viewpoint are used for inspection cost evaluation.  Since RRT* is a progressive optimization algorithm, we further modify the objective function into Equation (\ref{eq 3-12}) to determine the best parent node and achieve overall viewpoint optimization.

\begin{align}
    \textbf{Cost}\left(\mathbf{q}_{n e a r, i}, \mathbf{q}_{n e w}\right)&=\beta_{2} \frac{\sum_{j=1}^{n_{i, \mathbf{q}_{n e w}}^{\prime}} U_{i}\left(\mathcal{N}_{\mathbf{q}_{n e w}}^{\prime}(j)\right)}{n_{i, \mathbf{q}_{n e w}}^{\prime}+n_{i, \mathbf{q}_{n e a r}}^{\prime}}
    \notag\\
    &+\beta_{2} \frac{\sum_{k=1}^{n_{i, \mathbf{q}_{n e a r}}^{\prime}} U_{i}\left(\mathcal{N}_{\mathbf{q}_{n e a r}}^{\prime}(k)\right)}{n_{i, \mathbf{q}_{n e w}}^{\prime}+n_{i, \mathbf{q}_{n e a r}}^{\prime}}
    \notag\\
    &+\gamma_{2}\left(n_{i, \mathbf{q}_{n e w}}^{\prime}+n_{i, \mathbf{q}_{n e a r}}^{\prime}\right), i \in[1, \ldots, M],
    \label{eq 3-12}
\end{align}
where $M$ is the number of viewpoints in the neighbor domain of $\mathbf{q}_{new}$; $n_{i,\mathbf{q}_{new}}^{'}$ represents the number of first-time measured MPs in $\mathbf{q}_{new}$ when the parent node of $\mathbf{q}_{new}$ is the $i_{\text{th}}$ viewpoint in $\mathbf{q}_{near}$; $U_{i}(\mathcal{N}_{\mathbf{q}_{new}}^{'}(j))$ is the measurement uncertainty of the $j_{\text{th}}$ measured MP in $\mathbf{q}_{new}$ for the $i_{\text{th}}$ viewpoint; $n_{i,\mathbf{q}_{near}}^{'}$ represents the number of first-time measured MPs covered by the $i_{\text{th}}$ viewpoint in $\mathbf{q}_{near}$. The selection of the weights $\beta_{2}$ and $\gamma_{2}$ can be determined by the entropy method, so as to avoid the deviation caused by human factors.

Based on the inspection cost function, the viewpoint $\mathbf{q}_{min}$ with the smallest cost value in $\mathbf{q}_{near}$ is selected as the parent node of $\mathbf{q}_{new}$. The traditional RRT* algorithm generates the local path by reselecting the parent node and rewiring it after each iteration. In the proposed method, rather than considering the distance between viewpoints, the measurement uncertainty and the number of first-time-scanned MPs in the FOV are considered in the cost function. Therefore, we only need to reselect the parent node for the new viewpoint $\mathbf{q}_{new}$ to get the local optimal path without considering rewiring operations. The procedure of $\mathbf{ChooseParent}()$  is shown in Algorithm \ref{algorithm6}.

\begin{algorithm}[ht]
  \algsetup{linenosize=\tiny}
  \scriptsize
\SetAlgoLined
    Function: $\mathbf{q}_{min}=\textbf{ChooseParent}(\mathbf{q}_{near}, \mathbf{q}_{new}, \mathbf{q}_{nearest})$  \\
    $\mathbf{q}_{min}=\mathbf{q}_{nearest}$ \\
    $C_{min}=\textbf{Cost}(\mathbf{q}_{nearest}, \mathbf{q}_{new})$\\
    \For{$i=1:M$}{
    $C = \textbf{Cost}(\mathbf{q}_{near}, \mathbf{q}_{new})$\\
    \If{$C<C_{min}$}{
    $\mathbf{q}_{min} = \mathbf{q}_{near,i}$\\
    $C_{min} = C$
    }
    }
    \textbf{return} $\mathbf{q}_{min}$
\caption{Procedure to select parent node}
\label{algorithm6}
\end{algorithm}

\subsection{Sequential Planning of Sampled Viewpoints}
\label{sec 3.5}

Given the optimal set of viewpoints, the other sub-problem in CPP is to minimize the robot's motion time and ensure the inspection route is collision-free. This sub-problem can be solved by conducting local path planning and sequence planning sequentially. In the local path planning, we need to ensure the path between every pair of viewpoints (local path) is both collision-free and time-efficient. And then, sequence planning is to find the optimal order of the sampled viewpoints in $\mathbf{Q}$ to minimize the overall inspection time given the local path in the previous step. 

At first, we use the incremental sampling-based method to achieve a collision-free path between the two viewpoints. Specific steps refer to \cite{karaman2010optimal}. The robot moving time is then calculated, thereby establishing a collision-free time matrix $\mathbf{T}$, in which $t_{ij}$ represents the time consumption of the robot from the $i_\text{th}$ viewpoint to the $j_\text{th}$ viewpoint (spend over the collision-free local path and the rotation). $\mathbf{T}$ is a symmetric matrix with a diagonal of positive infinity. According to requirements in practice, the robot needs to return to its starting position after completing the scanning process of the last part. This problem can be formulated as a traveling salesman problem (TSP). In this paper, we consider the robot moving time between the adjacent viewpoints and the scanning time of the sensor in calculating the total time elapsed in the local path. The mathematical model to generate the optimal sequence of the sampled viewpoints is

\begin{align}
    \min \sum_{i=1}^{m} \sum_{j=1}^{m} t_{i j} x_{i j}+{t}_{m 0}+{t}_{01}+m t_{0},  \label{eq 3-9}\\
    \text { s.t. } \sum_{j=1}^{m} x_{i j}=1, i=1,2, \ldots, m, 
     \label{eq 3-10}\\
    \sum_{i=1}^{m} x_{i j}=1, j=1,2, \ldots, m,  \label{eq 3-15}
\end{align}
where $t_{01}$ represents the motion time when the robot starts from the initial position to reach the first viewpoint; $t_{m0}$ represents the time the robot takes to return to the initial position from the last viewpoint; $t_{0}$ represents the time the robot takes to complete inspection at the viewpoint. Without loss of generality, the inspection time for different types of MPs is assumed to be the same. $x_{ij}$ is a Boolean variable, when the path moves from the $i_\text{th}$ viewpoint to the $j_\text{th}$ viewpoint, $x_{ij}$ is $1$; Otherwise, $x_{ij}$ is $0$.

The TSP problem is a combinatorial optimization problem, and it is NP-hard. The methods to solve such problems include branch and bound mixed-integer linear programming and heuristic algorithms. The performance of the greedy algorithm, mixed-integer linear programming, simulated annealing algorithm, ant colony algorithm, and genetic algorithm were analyzed and compared \cite{liu2020optimal}. The results showed that the simulated annealing algorithm could obtain better results than other algorithms. We use the simulated annealing algorithm to obtain the optimal viewpoint sequence and generate the final collision-free inspection path. 

\section{Case study}
\label{sec4}
In order to evaluate the performance of the proposed method, a simulation test case with the inner panel of a car door and a real inspection experiment with a flange of the exhaust manifold are used. There are two evaluation criteria: the measurement uncertainty of key MPs and the inspection time of the path planning results. One cutting-edge CPP method, target sampling strategy \cite{glorieux2020coverage}, is used as a benchmark method for comparative analysis. The target sampling strategy was proposed for viewpoint sampling and optimization with an objective function. Its objective function considers the number of primitives within a candidate viewpoint and the distance of the inspection paths connecting the viewpoints.

\subsection{Simulation Test}
\label{sec 4.1}
\begin{figure}[ht]
    \includegraphics[width=\linewidth]{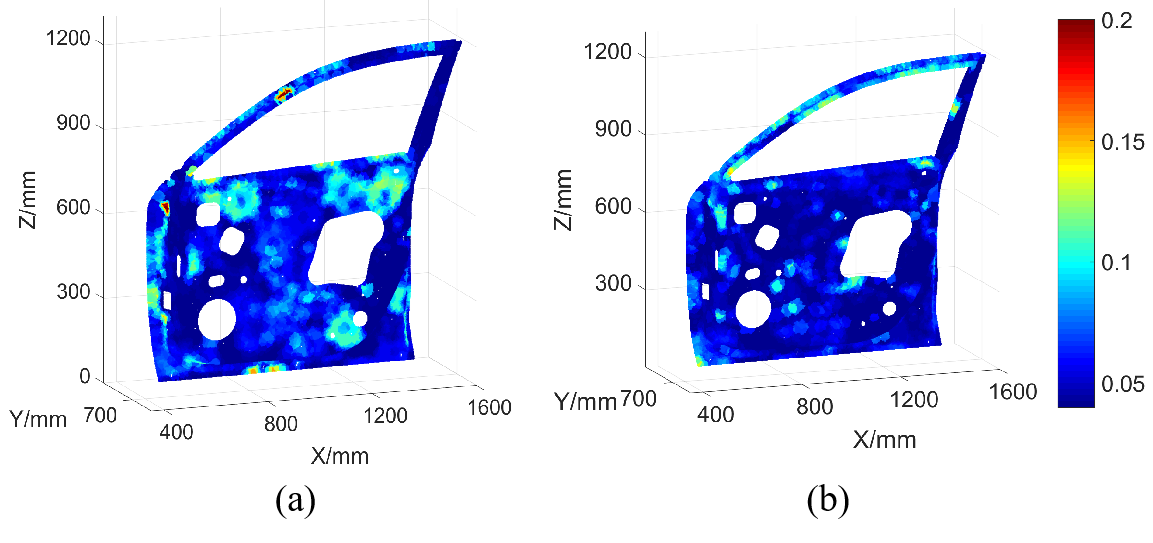}
    \centering
    \caption{The measurement uncertainty heatmap of the door inner panel (a) benchmark method; (b) proposed method}
    \label{Fig5}
\end{figure}

The coverage path planning of the inner panel in a car door is used to verify the performance of the proposed method for scanning large free-form surfaces. For the inner panel case, the number of voxels in the inner door panel is $2951$. The MPs, including the holes and trimmings, were extracted for dimensional verification of this sheet metal part. The number of the circular holes is $300$, and the number of the trimmings is $455$. Others are the non-critical points located on the panel surfaces. The surface roughness uncertainty of the part to be measured was set to be $0.01mm$. The tolerance for holes and slots is set as $\pm 0.5mm$. The tolerance for trimmings is $\pm 0.7mm$. The tolerance for the non-critical MPs is $\pm 1.0mm$. The amount, poses, and the sequential path of the viewpoints are planned based on the proposed method and the target sampling method, which is the benchmark method. Fig. \ref{Fig5} (a) and Fig. \ref{Fig5} (b) show the measurement uncertainty heatmap of the inner door panel based on the benchmark method and the proposed method, respectively. 

To demonstrate the measurement accuracy, the distributions of the measurement uncertainty based on the different sampling methods are shown in Table \ref{table1}. We can see that the range of measurement uncertainty using the proposed method is from $0.04mm$ to $0.13mm$, while the range of measurement uncertainty based on the target sampling is from $0.04mm$ to $0.19mm$. $88.82\%$ of MPs receive the measurement uncertainty within $[0.04, 0.07)$ using the proposed method, which is $8.64\%$ more than the results of the benchmark method. In the benchmark method, $4.47\%$ of MPs receive the measurement uncertainty within $[0.13, 0.19)$, while there is no MP receiving such large measurement uncertainty using the proposed method. Based on the optimized viewpoints, the measurement uncertainty of the inner door panel is reduced apparently, and the distribution of measurement uncertainty shrinks into a narrower range, thereby improving the quality inspection accuracy.

\begin{table}[ht]
\centering
\begin{threeparttable}[ht]
\caption{Results of the MU based on different viewpoint sampling strategies}
\newcommand{\tabincell}[2]{\begin{tabular}{@{}#1@{}}#2\end{tabular}}
{\begin{tabular}{ccc} 
\toprule
Range of MU & \tabincell{c}{The proposed method \\ No.(Ratio)} & \tabincell{c}{The target sampling method \\ No.(Ratio)} \\
\midrule
$[0.04,0.7)$ & 2621($88.82\%$) & 2366($80.18\%$)\\
$[0.07,0.1)$ & 198($6.71\%$) & 189($6.4\%$)\\
$[0.1,0.13)$ & 132($4.47\%$) & 264($8.95\%$)\\
$[0.13,0.16)$ & 0($0\%$) & 108($3.66\%$)\\
$[0.16,0.19)$ & 0($0\%$) & 24($0.81\%$)\\ 
\bottomrule
\end{tabular}}
\label{table1}
\begin{tablenotes}
\item MU: measurement uncertainty
\end{tablenotes}
\end{threeparttable}
\end{table}

\begin{table*}[ht]
\centering
\begin{threeparttable}[ht]
\caption{Comparative results of different viewpoint sampling methods}
{\begin{tabular}{ccccccccc} \toprule
& \multicolumn{2}{c}{Hole/slot} & \multicolumn{2}{c}{Trimming} & \multicolumn{2}{c}{Non-critical voxels} & \multirow{2}{4em}{Np. of VP.}& \multirow{2}{4em}{Inspection time ($s$)}\\ 
&Average MU &r &Average MU &r &Average MU &r & &\\
\midrule
The proposed method &0.049  &$100\%$ &0.053  &$100\%$ &0.058 &$100\%$ &206  &1199.97 \\
The target sampling method &0.0543  &$62.67\%$ &0.0589  &$91.21\%$ &0.0608 &$99.23\%$ &167  &976.33  \\
\bottomrule
\end{tabular}}
\label{table2}
\begin{tablenotes}
\item MU: measurement uncertainty
\end{tablenotes}
\end{threeparttable}
\end{table*}

Furthermore, the average measurement uncertainty of different types of MPs is calculated to evaluate the influence of the viewpoints sampling on the measurement accuracy. As shown in Table \ref{table2}, the average measurement uncertainty of the holes and slots based on the proposed method is $0.049mm$, while the average measurement uncertainty based on the benchmark method is $0.0543mm$, which is improved by $9.76\%$. The average measurement uncertainty of non-critical voxels is $0.058mm$ and is $4.61\%$ smaller than that based on the benchmark. Moreover, the ratio of the MPs or voxels that meet the uncertainty requirements is denoted by $r$. Based on the benchmark method, $62.67\%$, $91.21\%$, and $99.23\%$ of the holes/slots, trimmings, and non-critical voxels meet the uncertainty requirements, respectively, while the measurement uncertainties for all the MPs or the voxels are satisfied based on the proposed method.

\begin{figure}[ht]
    \includegraphics[width=\linewidth]{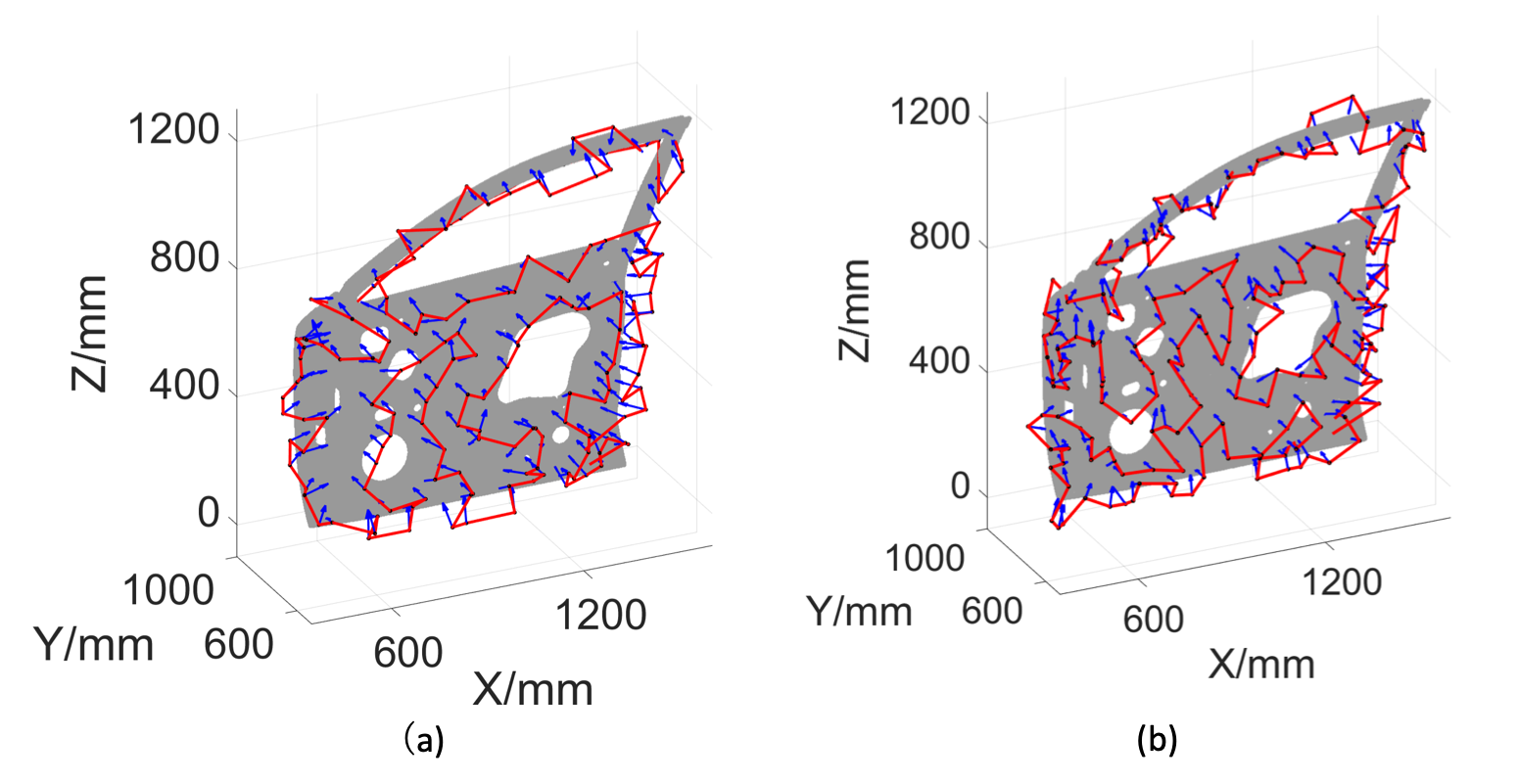}
    \centering
    \caption{Viewpoint sampling and CPP (a) benchmark method; (b) proposed method} 
    \label{Fig6}
\end{figure}

Fig. \ref{Fig6} (a) and Fig. \ref{Fig6} (b) showed the poses and sequences of the optimal viewpoints based on the benchmark method and the proposed method. The blue points and arrows denote the position and the incident angle of the sampled viewpoint. The red lines denote the optimal sequence of the sampled viewpoints. For the proposed method, the total number of viewpoints is $206$, and the robot scanning time is $1199.97s$. For the benchmark method, the number of viewpoints is $167$, and the robot scanning time is $976.33s$. The inspection time traversing the viewpoints was increased by $18.64\%$ compared with that via the benchmark. However, as shown in Table \ref{table2}, $37.33\%$ of holes/slots, $8.89\%$ of trimmings, and $0.77\%$ non-critical voxels located on the inner panel cannot meet the measurement uncertainty requirements when using the benchmark method. That is to say, the corresponding scanning data can not be considered accurate enough for product tolerance verification and process monitoring. 

Besides, the structured light gauges are generally used for off-line inspection, because it is difficult to measure the sheet parts for full coverage within a limited cycle time, for example, 60 seconds. Therefore, measurement accuracy is the most critical factor for off-line gauges. The enhanced RRT* algorithm can ensure the inspection accuracy for all MPs, with a slight increase in the inspection time. It can receive the optimal solution to the CPP problem.

\subsection[title]{Experimental Evaluation}
\begin{figure}[ht]
    \includegraphics[width=0.6\linewidth]{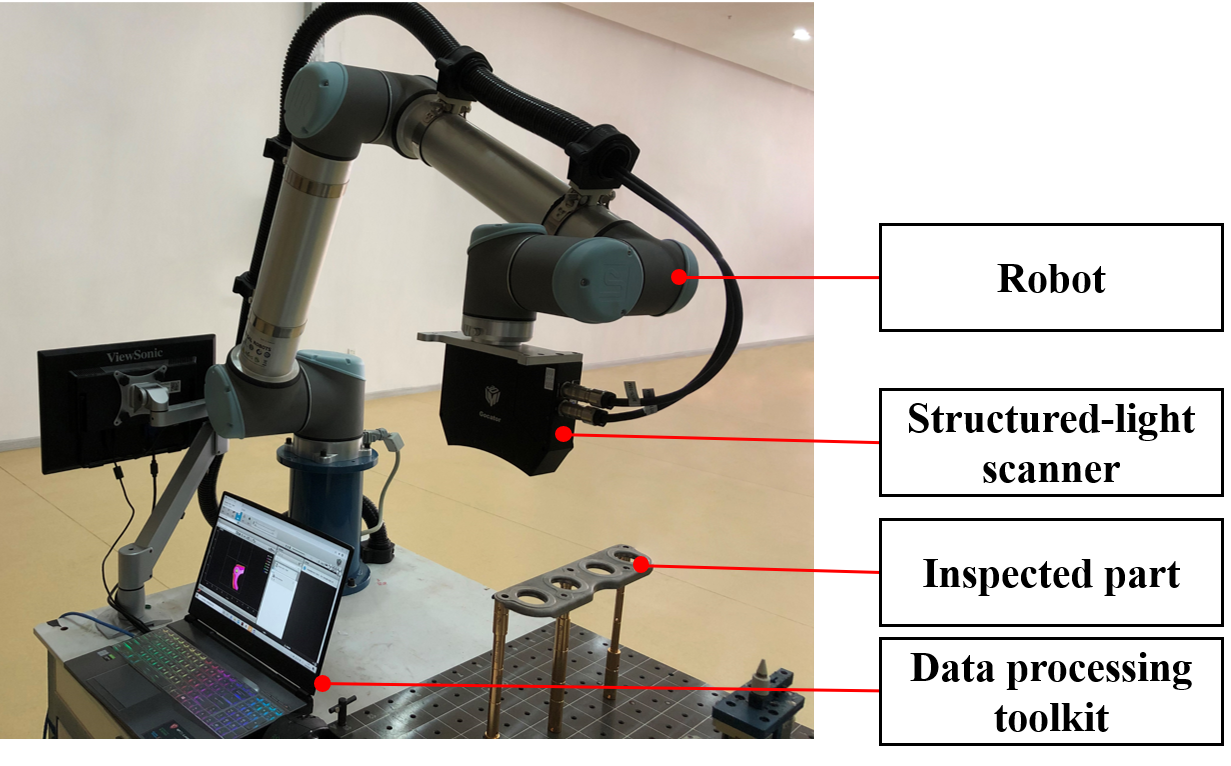}
    \centering
    \caption{Optical inspection system for the flange case} 
    \label{Fig8}
\end{figure}

A real case on the flange of the exhaust manifold is used for experimental evaluation. The scanning system is shown in Fig. \ref{Fig8}, which consists of the UR10 robot, an LMI Gocator 3210 line-structured light sensor, the scanning data processing toolkit, and the measured sheet metal part. Table \ref{table3} summarizes the sensor parameters of the optical probe. The key MPs located on this part mainly include nine holes and non-critical surface points. The tolerance of the holes is $\pm 0.5mm$, and the tolerance of non-critical surface points is $\pm 1.0mm$. We need to do CPP for full coverage of the flange part, and the measurement uncertainty needs to be satisfied during the path planning process. The measurement results of the key MPs, i.e., holes on the surface, based on different sampling strategies are compared. For evaluation purposes, we need to obtain the metrology data with ultra-higher precision. In this study, we use an ultra-high precision contact coordinate measurement machine (CMM). 

\begin{table}[ht]
\newcommand{\tabincell}[2]{\begin{tabular}{@{}#1@{}}#2\end{tabular}}
\centering
\caption{Sensor parameters}
{\begin{tabular}{ccccc}
\toprule
\tabincell{c}{DOF \\ (mm)} & \tabincell{c}{Near FOV \\ (mm)}&\tabincell{c}{Far FOV \\ (mm)}&\tabincell{c}{Scan depth \\ (mm)}&\tabincell{c}{Scan time/VP \\ (s)} \\ 
\midrule
100 & $60 \times 90$ & $90 \times 160$  & 250 & 5 \\
\bottomrule
\end{tabular}}
\label{table3}
\end{table}

\begin{figure}[htp]
\centering
    \includegraphics[width=\linewidth]{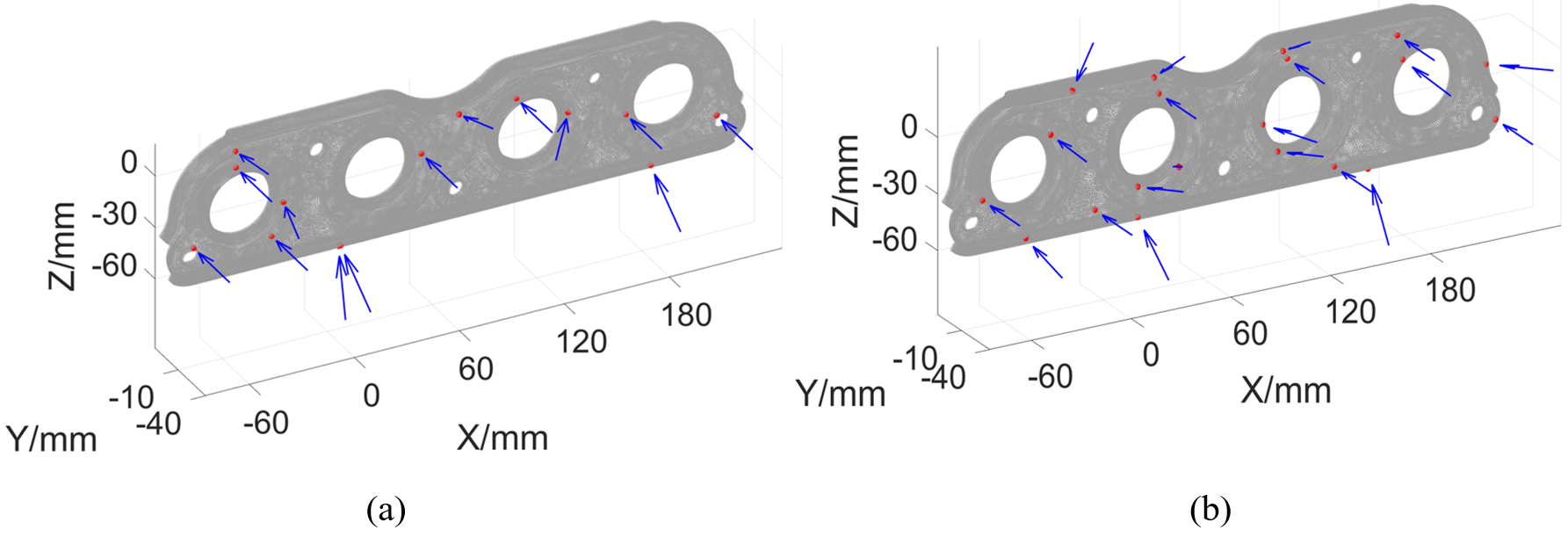}
    \caption{Optimal viewpoints of the flange (a)  benchmark method; (b)proposed method}
    \label{Fig9}
\end{figure}

The flange part is discretized into 787 voxels with grid processing. Based on the proposed enhanced RRT* algorithm, 20 optimal viewpoints are obtained, while 14 viewpoints are obtained based on the benchmark method. Fig. \ref{Fig9} (a) and Fig. \ref{Fig9} (b) show the optimized viewpoints and feasible incident angle based on these two methods, in which the red points represent the viewpoints and the blue arrows represent incident angles of the sampled viewpoints. The benchmark target sampling method focuses on minimizing the number of viewpoints and the scanning path. The proposed CPP method integrates the requirements of measurement uncertainty into the coverage planning to improve scanning accuracy. Although the number of viewpoints increases slightly, the proposed method can achieve more accurate shape reconstructing of holes and receive much higher precision in quality inspection for key MPs. At the same time, the measurement accuracy of non-critical MPs also improved with the proposed method.

\begin{figure}[htp]
\centering
    \includegraphics[width=\linewidth]{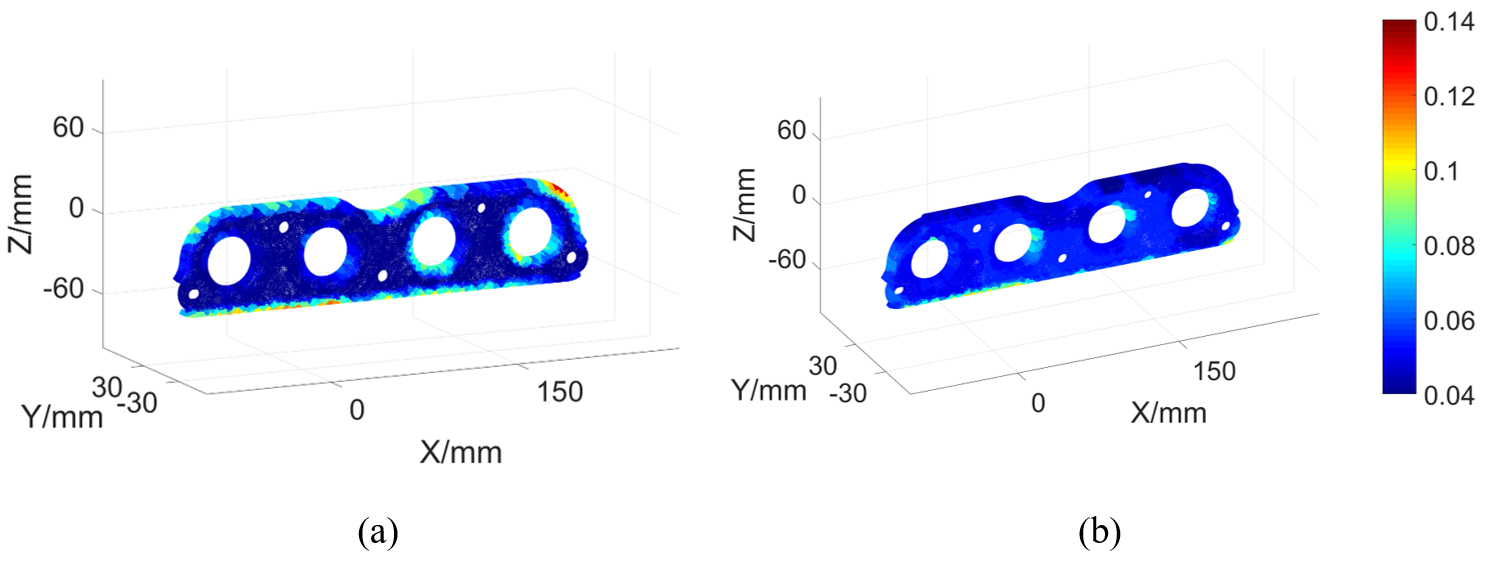}
    \caption{Measurement uncertainty heatmap of the flange (a) benchmark method; (b) proposed method}
    \label{Fig10}
\end{figure}

Fig. \ref{Fig10} (a) and Fig. \ref{Fig10} (b) show the measurement uncertainty heatmap based on the two CPP strategies. The measurement uncertainty of the flange scanning by the benchmark method is $0.04 \sim 0.14mm$, while the measurement uncertainty of the flange based on the enhanced RRT* algorithm is $0.04 \sim 0.11mm$. Table \ref{table4} shows the distribution of the measurement uncertainty under the optimized viewpoints. Based on the proposed method, the number of MPs with the measurement uncertainty within $[0.04,0.06)$ accounts for $93.27\%$ of the total MPs. While based on the benchmark method, the number of MPs with the measurement uncertainty within $[0.04,0.06)$ accounts for $79.28\%$. The number of MPs with measurement uncertainty within $[0.12,0.14)$ accounts for $5.6\%$ and $1.27\%$ for the benchmark method and the proposed method, respectively. In the case study, the proposed method can improve the measurement accuracy significantly.

\begin{table}[ht]
\centering
\begin{threeparttable}[ht]
\caption{The MU distribution with different viewpoints sampling methods}
\newcommand{\tabincell}[2]{\begin{tabular}{@{}#1@{}}#2\end{tabular}}
{\begin{tabular}{ccc} 
\toprule
Range of MU & \tabincell{c}{Target sampling \\ No.(Ratio)} & \tabincell{c}{The proposed method \\ No.(Ratio)} \\
\midrule
$[0.04,0.06)$ & 624($79.28\%$) & 734($93.27\%$)\\
$[0.06,0.08)$ & 59($7.5\%$) & 26($3.3\%$)\\
$[0.08,0.1)$ & 38($4.82\%$) & 17($2.16\%$)\\
$[0.1,0.12)$ & 22($2.8\%$) & 10($1.27\%$)\\
$[0.12,0.14)$ & 44($5.6\%$) & 0($0\%$)\\ 
\bottomrule
\end{tabular}}
\label{table4}
\begin{tablenotes}
\item MU: measurement uncertainty
\end{tablenotes}
\end{threeparttable}
\end{table}

\begin{table}[htp]
\newcommand{\tabincell}[2]{\begin{tabular}{@{}#1@{}}#2\end{tabular}}
\centering
\caption{Comparative results with the CMM measurements (mm)}
{\begin{tabular}{cccccc} \toprule
\tabincell{c}{Hole \\ No.} & \tabincell{c}{Target \\ sampling} & \tabincell{c}{Proposed \\ method}  & \tabincell{c}{CMM \\ Meas.} & $\Delta 1$ & $\Delta 2$\\ 
\midrule
1 & 4.484	&4.56	&4.358	&0.126	&0.202\\ 
2 & 19.905	&19.861	&19.505	&0.4	&0.356\\ 
3 & 5.193	&5.171	&5.134	&0.059	&0.037\\ 
4 & 19.864	&19.451	&19.556	&0.308	& 0.105\\ 
5 & 5.227	&5.169	&5.125	&0.102	&0.044\\ 
6 & 19.888	&19.453	&19.535	&0.353	&0.082\\ 
7 & 4.429	&4.4	&4.358	&0.071	&0.0425\\ 
8 & 19.976	&19.451	&19.603	&0.373	&0.152\\ 
9 & 5.254	&5.25 	&5.149	&0.105	&0.101\\
\rowcolor{lightgray} \tabincell{c}{Average \\ devi.} & 	& 	&	&0.211	&0.125\\
\bottomrule
\end{tabular}}

\label{table5}
\end{table}

Furthermore, the data processing software of the Gocator3210 line-structured light sensor is used to fit the radius of the holes based on the two methods. And the radius results are compared with the results from the high-precision CMM with a measurement precision of $\pm 0.007mm$. The comparison is shown in Table \ref{table5}. $\Delta 1$ represented the radius deviations of the holes between the scanning results of the benchmark method and CMM measurement data, while $\Delta 2$ represented the deviations between the scanning results of the proposed method and CMM measurement data. The maximum radius deviation based on the benchmark method is $0.4mm$, and the minimum deviation is $0.059mm$. Based on the proposed algorithm, the maximum radius deviation is $0.356mm$, and the minimum deviation is only $0.037mm$. According to Table \ref{table5}, the average deviation for the hole inspection based on the proposed method is $40.88\%$ smaller than that of the benchmark method. Therefore, with the same hole fitting algorithm, more accurate point cloud data is obtained with the proposed viewpoint sampling strategy.

\section{Conclusion}\label{sec5}
By incorporating the measurement uncertainty, a novel viewpoint sampling and sequential planning algorithm is proposed for CPP and quality inspection of free-form surfaces. Based on the tolerance specifications of MPs on the parts, the feasible ranges of measurement uncertainty are calculated, and the initial viewpoint set is further generated considering the feasible ranges of measurement uncertainty and visibility of MPs. In terms of viewpoint optimization, the enhanced RRT* algorithm is proposed by integrating the measurement uncertainty and the number of viewpoints into the inspection cost evaluation. At last, the viewpoint sequential planning is conducted based on the simulated annealing algorithm. Both simulation and case studies are performed to evaluate the effectiveness of the proposed method. Results show the scanning errors are reduced significantly compared to the benchmark method. Overall, the proposed method can achieve higher precision in quality inspection. 

\bibliographystyle{IEEEtran}

\bibliography{cpp_trans}
\vskip -3\baselineskip plus -1fil






\appendix

\section{Appendix A}

\textbf{Appendix A: Functions used in algorithm 1}

$\textbf{Boundingboxset}()$ is the function to evaluate whether the MP is in the inspection region $\Omega$, according to the FOV established in Section \ref{sec3.2}.

$\textbf{Robotaccessibility}()$ is a function to evaluate the accessibility of a robot to the viewpoint $\mathbf{q}_{i}$. Given the pose of the viewpoint $\mathbf{q}_i$, each joint angle of the robot can be calculated based on the inverse kinematics of the robot \cite{efac4aac9f874f95850c6106d0ff8377}. If the joint angle of the robot exists, the viewpoint is reachable from the robot(marked as $1$), otherwise, it is unreachable (marked as $0$).

$\textbf{Collision}()$ is a function to evaluate whether the robot collides with the static environment when inspecting the viewpoint $\mathbf{q}_{i}$ \cite{liu2020optimal}. If the collision occurs, it is $1$, otherwise, it is $0$.

\vspace{10mm}

\textbf{Appendix B: Functions used in algorithm 2}

$\textbf{CalSampling}()$: The initial viewpoint set determined in Section \ref{sec3.2} is calculated.

$\textbf{Sample}()$: randomly select viewpoint $\mathbf{q}_{rand}$.

$\textbf{Extend}()$: Select the viewpoint from $\mathcal{T}$ that is closest to the line $\overline{\mathbf{q}_{rand},\mathbf{q}_{nearest}}$, and the details are shown in Algorithm \ref{algorithm3}.

$\textbf{FindbestNeighbor}()$: Search the viewpoint from the set $\mathbf{Q}$ that has the minimum cost (defined in Equation (\ref{eq 3-12})) to  $\mathbf{q}_{new}$. The details are shown in Algorithm \ref{algorithm5}.

$\textbf{FeasiblePoint}()$: If the number of MPs contained in this candidate viewpoint $\mathbf{q}_{new}$ is greater than 0 and the robot does not collide, then $\textbf{FeasiblePoint}()=1$, otherwise $\textbf{FeasiblePoint}()=0$.

$\textbf{Neighbors}()$: The subset of viewpoints in $\mathbf{Q}$ which locate within a certain distance to  $\mathbf{q}_{new}$.

$\textbf{ChooseParent}()$: Select the viewpoint $\mathbf{q}_{min}$ with the smallest cost in $\mathbf{q}_{near}$ as the parent node of $\mathbf{q}_{new}$.

\vspace{10mm}

\textbf{Appendix C: Summary Table of Notations}

\begin{table*}[ht]
\centering
\begin{threeparttable}[ht]
\caption{Summary of Notations}
{\begin{tabular}{cc} \toprule
Notation & Explanation\\ 
\midrule
$k$&coverage factor\\

$U_{ad}$&maximum measurement uncertainty\\

$U_{sen}$&uncertainty of the sensor\\

$U_{mat}$&uncertainty of the material\\

$U_{rot}$&uncertainty of the robot\\

$\mathbf{n}_{l}$&direction of the sensor\\

$\mathbf{q}_{i}$&$i_{\text{th}}$ viewpoint\\

$\mathbf{P}$&full set of MPs\\

$n_{p}$&vector direction of the $p_{\text{th}}$ MP\\

$\alpha_p$&maximum feasible incident angle of the $p_{\text{th}}$ MP\\

$\Omega$ & Feasible inspection area located under FOV\\

$\mathbf{S}$&subset of MPs contained in $\Omega$\\

$m$&total number of selected viewpoints\\

$m_{\mathbf{S}}$&the number of MPs contained in the set $\mathbf{S}$\\

$m_{un}$&number of viewpoints in $\mathcal{T}_{unselect}$\\

$m_{\mathbf{Q}}$&the number of viewpoints contained in $\mathbf{Q}$\\

$\mathcal{N}_{\mathbf{q}_{i}}^{'}$&set of MPs newly included in viewpoint $\mathbf{q}_{i}$\\

$\mathcal{N}_{\mathbf{q}_{new}}^{\prime}$&set of measured points contained in viewpoint $\mathbf{q}_{new}$\\

$U(\mathcal{N}_{\mathbf{q}_{i}}^{'}(j))$&measurement uncertainty of the $j_{\text{th}}$ MP in $\mathcal{N}_{q_i}^{'}$\\

$U_{i}(\mathcal{N}_{\mathbf{q}_{new}}^{'}(j))$&measurement uncertainty of the $j_{\text{th}}$ measured MP in $\mathcal{N}_{\mathbf{q}_{new}}^{\prime}$ under the $i_{\text{th}}$ viewpoint\\ 

$\mathbf{G}(\mathbf{q}_{i})$&geometric space occupied by the inspection system at the viewpoint $\mathbf{q}_{i}$\\

$\mathbf{V}$&geometric space occupied by the inspected surface system\\

$\beta_{1}$,$\gamma_{1}$,$\beta_{2}$,$\gamma_{2}$&weight factor\\

$\mathbf{Q}$&sampled optimal viewpoint set\\

$\mathbf{P^{\prime}}$&set of MPs contained in $\mathbf{Q}$\\

$\mathbf{P}_{\mathbf{q}_{new}}$ & $\mathbf{P}_{\mathbf{q}_{new}}$ is determined by tracing back from the current root node $\mathbf{q}_{new}$ to the initial viewpoint recursively\\

$\mathcal{G}$ & a directed tree graph\\

$\mathcal{T}$&candidate viewpoint set\\

$env$&static environment including the measured part and tooling, etc.\\

$\mathbf{q}_{init}$&initial position of robot end-effector\\

$\mathbf{q}_{rand}$&randomly generated viewpoint\\

$\mathbf{q}_{nearest}$&closest viewpoint to $\mathbf{q}_{new}$ in $\mathbf{Q}$\\

$\mathbf{q}_{near}$&viewpoint in $\mathbf{Q}$ whose distance from the viewpoint  to $\mathbf{q}_{new}$ is less than the threshold\\

$\mathbf{q}_{new}$&newly determined viewpoint\\

$\mathbf{q}_{best}$ & the last optimal viewpoint selected.\\

$\mathbf{q}_{min}$&viewpoint closest to $\mathbf{q}_{new}$ in $\mathcal{T}$\\

$\mathcal{T}_{unselect}$&viewpoint set that is not selected in $\mathcal{T}$\\

$D_{0}$&shortest distance from $\mathbf{q}_{new}$ to the line constructed by $\mathbf{q}_{rand}$ and $\mathbf{q}_{nearest}$\\

$\epsilon$ & distance threshold\\

$M$&the number of viewpoints in the neighbor domain of $\mathbf{q}_{new}$\\

$N$&the number of metal parts divided into voxels\\

$n_{i,\mathbf{q}_{new}}^{'}$&the number of first-time measured MPs in $\mathbf{q}_{new}$ when its parent node is the $i_{\text{th}}$ viewpoint in $\mathbf{q}_{near}$\\

$n_{i,\mathbf{q}_{near}}^{'}$&the number of first-time measured MPs covered by the $i_{\text{th}}$ viewpoint in $\mathbf{q}_{near}$\\

$t_{ij}$&total time of movement and rotation of the robot from the $i_\text{th}$ viewpoint to the $j_\text{th}$ viewpoint\\

$t_{01}$&motion time when the robot starts from the initial position to reach the first viewpoint\\

$t_{m0}$&the time for the robot to return to the initial position from the last viewpoint\\

$t_{0}$&inspection time of the robot to complete a viewpoint\\

$x_{ij}$&Boolean variable\\

\bottomrule
\end{tabular}}
\label{table6}
\end{threeparttable}
\end{table*}

\end{document}